\documentclass[letterpaper, 10 pt, conference]{ieeeconf}  

\IEEEoverridecommandlockouts                              

\usepackage[nocompress]{cite}
\usepackage{amsmath,amssymb,amsfonts}
\usepackage{algorithm}
\usepackage{algpseudocode}
\usepackage{graphicx}
\usepackage{textcomp}
\usepackage{xcolor}
\usepackage{subcaption}
\usepackage{booktabs} 
\usepackage{amssymb}  
\usepackage{multirow} 
\usepackage{booktabs} 
\usepackage{makecell} 
\usepackage{tabularx} 
\usepackage{censor}
\usepackage{balance}

\usepackage[capitalize]{cleveref}
\title{\LARGE \bf
Sim-to-Real Aware End-to-End Learning Environment for Micromobility
}

\author{Shouma Amano$^{1}$ and Takuya Azumi$^{2}$
\thanks{$^{1}$Saitama University
        {\tt\small amano.s.388@ms.saitama-u.ac.jp}}%
\thanks{$^{2}$Academic Association (Graduate School of Science and Engineering) Saitama University}%
}
\begin{document}

\maketitle
\thispagestyle{empty}
\pagestyle{empty}

\begin{abstract}

While end-to-end autonomous driving systems show promise, their application to micromobility vehicles is hindered by simulators failing to capture specific kinematics, such as differential drives and omni-wheels.
This paper proposes a sim-to-real-aware, vehicle-specific end-to-end learning environment for the WHILL Model CR on AWSIM and ROS~2.
To minimize the sim-to-real gap, physical parameters are optimized via Bayesian optimization using real-world data, reducing trajectory errors across various driving scenarios.
Additionally, this study introduces a synchronized architecture tailored for the stable training of world model-based agents.
An end-to-end policy trained with DreamerV3 exhibited learning progress and achieved task completion in a simulated obstacle avoidance setting.
Furthermore, this policy demonstrated direct sim-to-real transfer to the physical vehicle, enabling the vehicle to navigate around a cardboard box in a real-world corridor replica without fine-tuning.
This paper provides a practical foundation for sim-to-real micromobility policy studies.


\end{abstract}

\section{INTRODUCTION}\label{sec:Introduction}

Autonomous driving systems~\cite{kato2018autoware, tajima2024ros} are expanding to micromobility vehicles, such as delivery robots and electric wheelchairs.
This technology enables inclusive, efficient, and sustainable urban environments through novel transportation and last-mile logistics. 
However, integrating these vehicles requires systems capable of complex interactions with pedestrians and dynamic obstacles.
Traditional modular systems face challenges in these environments, particularly suboptimal module coordination and error propagation from perception to planning.
End-to-end (E2E)~\cite{yu2025end, chen2024end, hwang2024emma} approaches address these challenges by utilizing a single, unified network mapping raw sensor data to driving commands. Furthermore, the growing use of world models~\cite{feng2025survey,hu2023gaia,guan2024world,wu2023daydreamer} enables agents to build an internal representation of their environment and simulate future outcomes, allowing for proactive, long-term planning.

While these learning-based methods have largely focused on conventional automobiles, applying them to micromobility vehicles presents two barriers. First, real-world testing on unpredictable sidewalks is costly and risky, making simulation an essential tool. Second, current mainstream autonomous driving simulators are insufficient, as they are primarily designed to replicate the dynamics and traffic flow of cars. 
Standard simulators often fail to capture the complex kinematics inherent to micromobility vehicles, such as differential drive systems and omni-wheel behaviors. This physical discrepancy creates a sim-to-real gap, which hinders the deployment of E2E driving policies.

To overcome these barriers, this study proposes an E2E learning environment tailored for the WHILL Model CR using AWSIM~\cite{ito2025d} and ROS~2. To minimize the sim-to-real gap, a system identification framework utilizing Bayesian optimization is implemented to tune physical parameters based on real-world driving data. Furthermore, to support E2E learning, this study introduces a layered architecture featuring a hybrid synchronization protocol. This protocol bridges the asynchronous ROS~2 data streaming with episodic reinforcement learning requirements, ensuring stable training for world model-based agents. This E2E learning environment enables the direct sim-to-real transfer of the learned policy to the physical vehicle without additional real-world training.

The main contributions of this paper are as follows:

\begin{itemize}

\item Minimizing the sim-to-real gap in trajectory and yaw angle responses by optimizing physical parameters based on real-world driving data.
\item Establishing an E2E learning infrastructure for micromobility vehicles using ROS~2 and AWSIM, demonstrated to support E2E driving policy training.
\item Demonstrating direct sim-to-real transfer of an E2E driving policy in the reported corridor scenario, enabling collision-free obstacle avoidance without updating the network weights.

\end{itemize}

Based on these contributions, this study addresses the following research questions:
\begin{itemize}

\item RQ1: To what extent can the sim-to-real gap in trajectory and yaw angle be minimized by optimizing physical parameters based on real-world driving data?

\item RQ2: In the proposed environment, can a world model-based E2E policy (DreamerV3) trained solely on RGB images achieve stable obstacle avoidance performance, as measured by reward curve?

\item RQ3: Can the vision-based E2E policy, trained in the proposed environment, achieve collision-free obstacle avoidance on a physical vehicle without network weight fine-tuning?

\end{itemize}

The paper is organized as follows: Section~\ref{sec:System model} describes the system model, and Section~\ref{sec:Design and implementation} details the proposed environment. Section~\ref{sec:evaluation} reports the evaluation, Section~\ref{sec:Related work} discusses related work, and Section~\ref{sec:conclusions} concludes.

\begin{figure}[t]
    \centerline{\includegraphics[width=\linewidth]{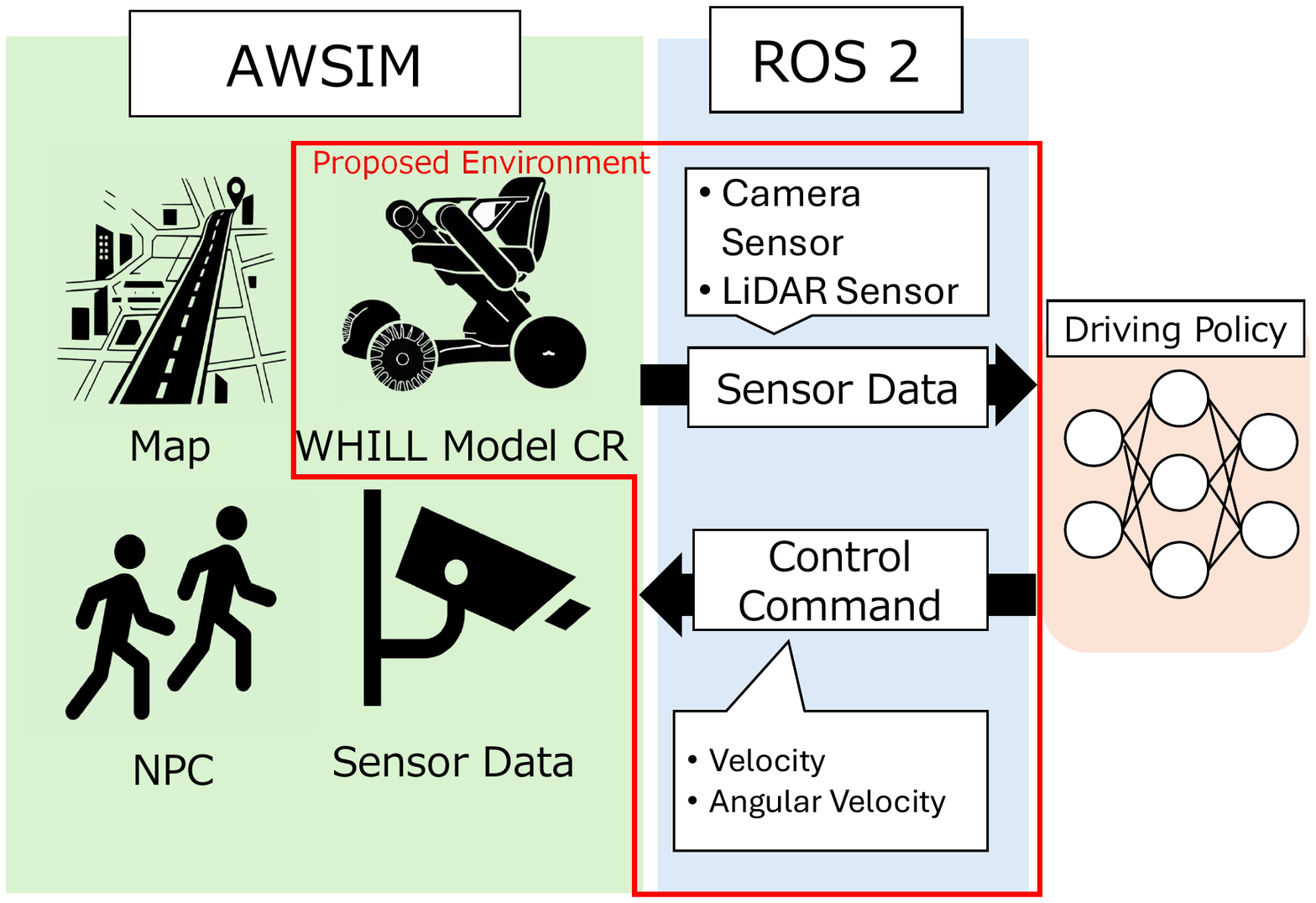}}
    \caption{System Model.}\label{fig1}
\end{figure}

\section{SYSTEM MODEL}\label{sec:System model}

The proposed system model integrates a simulator, a micromobility vehicle, and a world model-based driving policy via middleware, as depicted in Fig.~\ref{fig1}.

\subsection{AWSIM}\label{ssub:AWSIM}
AWSIM is a digital twin-oriented autonomous driving simulator designed to replicate real-world environments. In this study, a detailed map of Nishi-Shinjuku, Tokyo, is utilized and populated with Non-Player Characters (NPCs) to collect the multimodal sensor data (camera and LiDAR) required for training the agent.

\subsection{WHILL Model CR}\label{ssub:WHILL}
The target vehicle is the WHILL Model CR, a micromobility vehicle characterized by a differential drive system for propulsion and specialized front omni-wheels. This combination enables distinct kinematic behaviors, such as zero-radius turns with minimal lateral friction, which influence the range of actions the driving policy must learn.

\subsection{ROS~2}\label{ssub:ROS~2}

ROS~2~\cite{maruyama2016exploring} bridges the AWSIM simulator and the learning agent. This middleware handles the asynchronous transmission of sensor data to the agent and relays velocity control commands back to the vehicle.

\subsection{Driving Policy}\label{ssub:world model}
The driving policy employs DreamerV3~\cite{hafner2023mastering}, a world model-based reinforcement learning algorithm. DreamerV3 was selected because the reported task requires learning continuous driving commands from low-resolution RGB observations under a limited number of simulator interactions, where model-based latent dynamics can improve sample efficiency and temporal prediction compared with purely reactive policies. This choice is also consistent with prior physical robot learning studies using world models~\cite{wu2023daydreamer}. Within the proposed environment, this E2E policy functions as a neural controller that maps raw sensor observations directly to driving commands, adapting to the specific kinematics of the micromobility vehicle.

\section{DESIGN \& IMPLEMENTATION}\label{sec:Design and implementation}

\begin{figure*}[t]
    \centering
    \includegraphics[width=\textwidth]{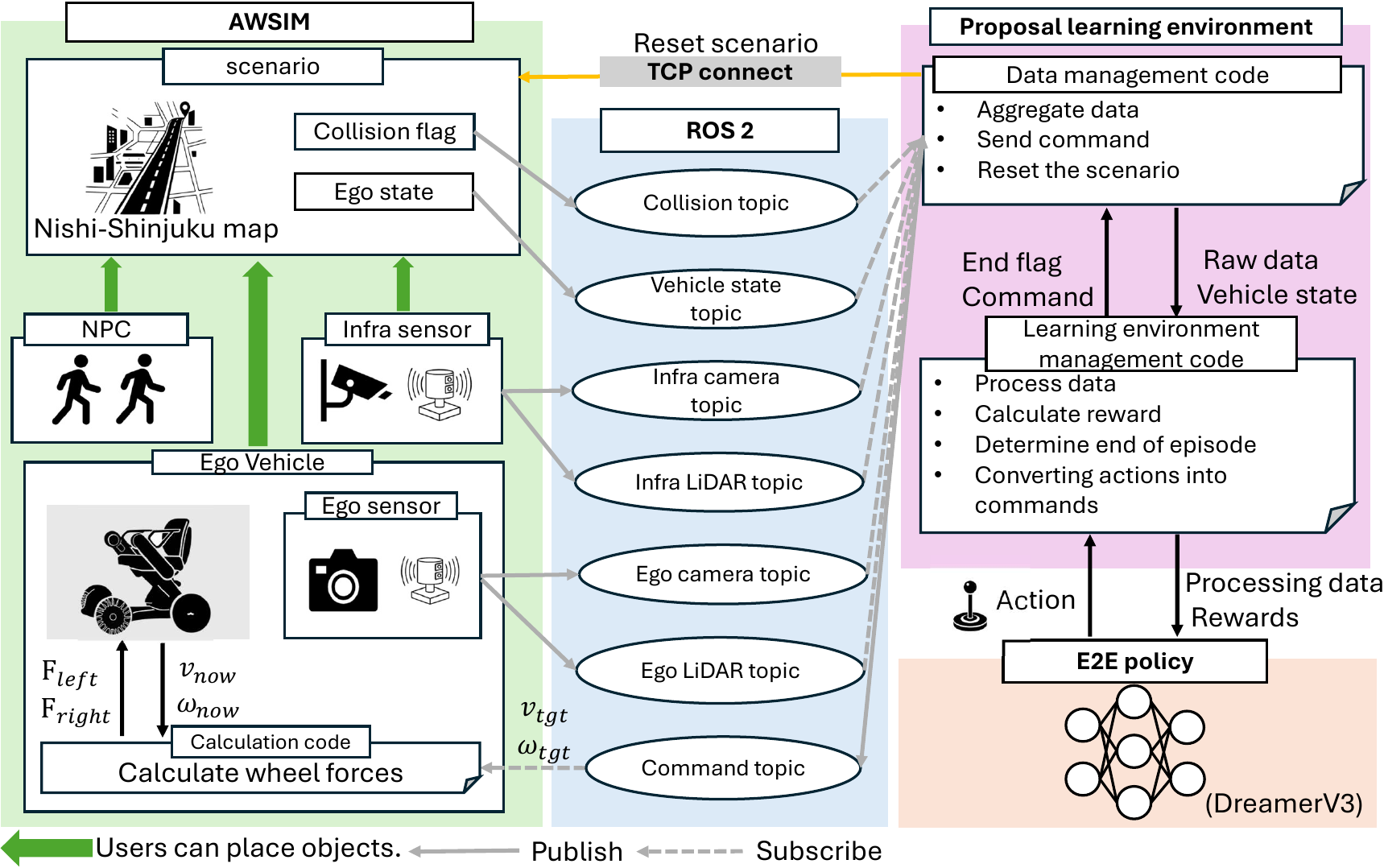}
    \caption{Detailed System Model}\label{fig3}
\end{figure*}

The proposed integrated E2E learning environment architecture utilizes AWSIM and ROS~2 to enable world model-based reinforcement learning for micromobility vehicles.
The WHILL Model CR is modeled within this environment to provide a foundation for training and deploying E2E driving policies, as illustrated in Fig.~\ref{fig3}.

\subsection{E2E Learning Environment and Control Interface}\label{E2E learning environment and Control}
\begin{figure}[t]
    \centerline{\includegraphics[width=0.8\linewidth]{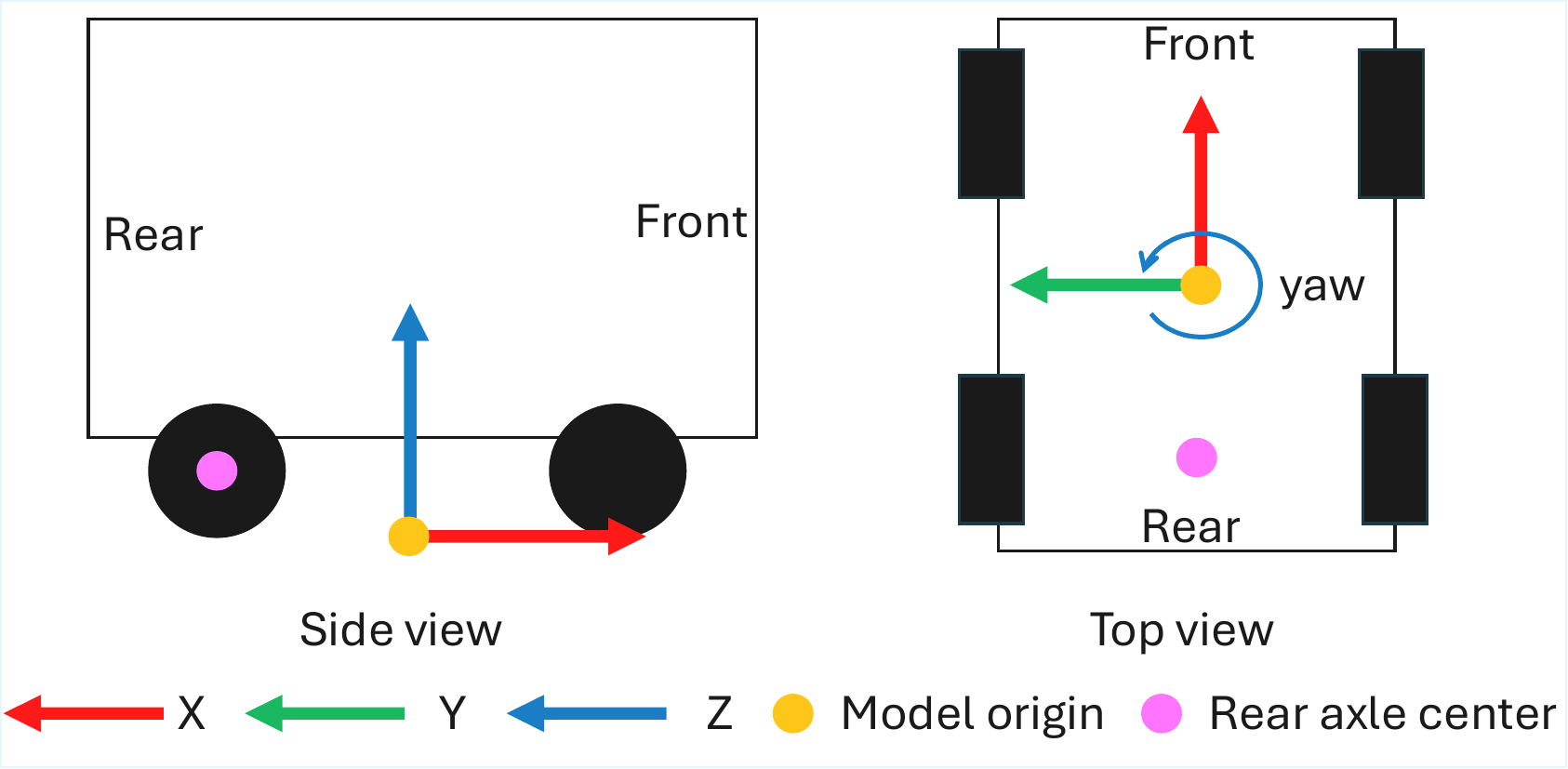}}
    \caption{Definition of the vehicle coordinate system and reference points.}\label{coordinate}
    
\end{figure}

A 3D model of the WHILL Model CR is integrated into the AWSIM environment. The simulation utilizes a map of Nishi-Shinjuku to collect multimodal sensor data in wheelchair-accessible areas. To bridge the internal physics engine with the ROS~2 ecosystem, the coordinate systems and kinematic reference points are defined, as illustrated in Fig.~\ref{coordinate} and Table~\ref{tab:coord_def}. The vehicle is controlled via standard ROS~2 \texttt{geometry\_msgs/Twist} messages published to the \texttt{cmd\_vel} topic, defining target linear ($v$) and angular ($\omega$) velocities. Furthermore, the proposed environment supports the evaluation of direct sim-to-real transfer without network weight fine-tuning. For this study, a virtual replica of a university laboratory corridor was constructed. This scenario incorporates 3D models of task-specific objects, such as a cardboard box and a goal marker, to evaluate visual obstacle avoidance.

\begin{table}[tb]
    \centering
    \caption{Definitions of coordinate systems and reference frames}
    \label{tab:coord_def}
    \renewcommand{\arraystretch}{1.2}
    \footnotesize 
    \resizebox{\columnwidth}{!}{%
    \begin{tabular}{| l | c | p{3.5cm} |} 
        \hline
        \textbf{Category} & \textbf{Symbol} & \textbf{Definition \& Description} \\
        \hline
        \textbf{Coordinate System} & $\Sigma_{base}$ & \textbf{Right-Handed (ROS Std.)} \newline
        Defined at the Model Origin. \newline $+x$: Forward, $+y$: Left, $+z$: Up. \\
        \hline
        \textbf{CoM Offsets} & $CoM_{x,y,z}$ & \textbf{Relative to $\Sigma_{base}$} \newline
        Offsets from the Model Origin. \newline $CoM_x$: Front, $CoM_y$: Left, $CoM_z$: Up. \\
        \hline
        \textbf{Control Inputs} & $v, \omega$ & \textbf{Ref. at Rear Axle Center} \newline
        $v$: Linear velocity ($+x$). \newline $\omega$: Angular velocity (yaw). \\
        \hline
        \textbf{Evaluation} & $\Sigma_{eval}$ & \textbf{Odometry Frame} \newline
        Tracks the 2D pose of the Rear Axle Center relative to the initial starting position. \\
        \hline
    \end{tabular}
    } 
\end{table}

\subsection{Vehicle Modeling and Parameter Optimization}\label{tires}
To bridge logical velocity commands and the physics engine, a force-based control algorithm designed for the differential drive system of the WHILL Model CR is utilized.
Target linear ($v_{tgt}$) and angular ($\omega_{tgt}$) velocities are decomposed into left and right wheel velocities using standard inverse kinematics. Independent PID controllers then calculate the required acceleration for each wheel to minimize velocity tracking errors. Finally, these accelerations are mapped to actual tractive forces ($F_{left}, F_{right}$) applied to the simulated wheels, assuming a simplified mass distribution.

A system identification problem is formulated to minimize the sim-to-real gap.
This optimization focuses on a parameter set $\theta$ (including friction coefficients and control gains) to minimize the trajectory tracking error between open-loop executions on the physical vehicle and the simulation.
The objective function $J(\theta)$ is the sum of the Root Mean Square Errors (RMSE) across a set of driving scenarios $S$:
\begin{equation}
J(\theta)=\sum_{k\in S}\sqrt{\frac{1}{T_{k}}\sum_{t=1}^{T_{k}}||\mathbf{p}_{sim}^{(k)}(t,\theta)-\mathbf{p}_{real}^{(k)}(t)||^{2}}   
\end{equation}
where $T_{k}$ is the total steps in scenario $k$, and $S$ comprises five predefined maneuvers (Straight, Turn, Circle, S-curve, and Figure-eight) designed to capture the longitudinal and non-linear lateral dynamics of the differential drive system. Here, $\mathbf{p}_{sim}$ represents the ground-truth 2D position of the rear axle center of the vehicle (Fig.~\ref{coordinate}), extracted from the global coordinates of the simulator. Correspondingly, $\mathbf{p}_{real}$ denotes the same rear axle center position of the physical vehicle, estimated via wheel odometry. For a kinematic comparison, both trajectories are aligned to a common origin at the start of each trial.
The optimized parameter set $\theta$ was determined through 500 trials of Bayesian optimization (TPE~\cite{bergstra2011algorithms}) conducted via Optuna~\cite{akiba2019optuna}.
The initial and optimized values for the target parameters are presented in Table~\ref{table:scenario_params}.

\begin{table}[tb]
\centering
\caption{Identified system parameters for the vehicle dynamics model}
\label{table:scenario_params}
\renewcommand{\arraystretch}{1.1}
\setlength{\tabcolsep}{2pt}
\scriptsize
\begin{tabularx}{\columnwidth}{| c | X | c | c |}
\hline
\textbf{Symbol} & \textbf{Description (Unit)} & \textbf{Initial Values} & \textbf{Optimized Values$^*$} \\ \hline
\multicolumn{4}{|l|}{\textbf{Physics (Skidding \& Slip)}} \\ \hline
$c_{rear}$ & Lateral damping coefficient (rear) [$1/s$] & 1.000 & 34.79 \\ \hline
$c_{front}$ & Lateral damping coefficient (front) [$1/s$] & 0.1000 & 0.4056 \\ \hline
$\eta$ & Forward slip multiplier [$-$] & 1.000 & 0.8400 \\ \hline

\multicolumn{4}{|l|}{\textbf{PID Control}} \\ \hline
$K_p$ & Proportional gain [$1/s$] & 10.00 & 27.70 \\ \hline
$K_i$ & Integral gain [$1/s^2$] & 0.0000 & 0.4025 \\ \hline
$K_d$ & Derivative gain [$-$] & 0.0000 & 0.7384 \\ \hline
$I_{clamp}$ & Anti-windup limit [$m$] & 0.5000 & 1.431 \\ \hline

\multicolumn{4}{|l|}{\textbf{Resistance}} \\ \hline
$C_d$ & Air drag coefficient [$-$] & 0.2000 & 1.132 \\ \hline
$\mu_r$ & Rolling resistance coefficient [$-$] & 0.002000 & 0.003300 \\ \hline
$A$ & Frontal area [$m^2$] & 0.8000 & 1.062 \\ \hline

\multicolumn{4}{|l|}{\textbf{CoM Offsets}} \\ \hline
$\text{CoM}_x$ & Longitudinal [$m$] & 0.0000 & $-0.2654$ \\ \hline
$\text{CoM}_y$ & Lateral [$m$] & 0.0000 & 0.004400 \\ \hline
$\text{CoM}_z$ & Vertical [$m$] & 0.0000 & 0.1215 \\ \hline
\end{tabularx}

\vspace{0.5ex}
\raggedright 
\scriptsize $^*$Optimized values were obtained using training data from four scenarios: Straight, Turn, Circle, and S-curve.
\end{table}

The identified parameters should be interpreted as simulator-level corrections rather than independently measured physical constants, with the dominant mismatch attributed to lateral slip and yaw-coupled motion.

\subsection{E2E Learning and Deployment Architecture}
The architecture enables DreamerV3 training and deployment to the vehicle.
An environment wrapper converts raw sensor streams into model-ready tensors (e.g., $64 \times 64$ images).
Furthermore, this wrapper bridges the action space by mapping the normalized continuous actions output by the DreamerV3 agent ($a_t = [a_v, a_\omega] \in [-1, 1]^2$) to standard \texttt{geometry\_msgs/Twist} commands.
These normalized values are linearly scaled to predefined velocity limits ($v_{max}$ and $\omega_{max}$) tailored to the specific experimental scenario.
This transformation ensures that the generated commands remain within kinematically feasible and safe boundaries for the vehicle.
Because both the AWSIM and the physical WHILL Model CR share the identical ROS~2 interface, the trained E2E policy is deployed directly to the real world without code modification.

A primary challenge in coupling asynchronous ROS~2 middleware with episodic reinforcement learning is synchronization.
To prevent state mismatches during environment resets, a Hybrid Communication Protocol is implemented: high-bandwidth sensor data streams continuously via ROS~2, while state-management signals (e.g., episode resets) are transmitted via a synchronous TCP connection.
To ensure the low-latency observation required for policy stability, the synchronization of asynchronous ROS~2 messages follows the specifications defined in Table~\ref{tab:ros2_specs}.
Since the simulation physics and the inference process of the agent operate asynchronously, the subscriber queue size for high-bandwidth sensors is limited to a depth of 1. This drop-oldest policy ensures that the agent always retrieves the most recent frame available at the moment of inference, discarding any stale data accumulated due to processing jitter.

In the proposed environment, the decision step of the Markov Decision Process (MDP) is defined by the arrival of a new camera frame.
Let $\Delta t_{dec}$ denote the target inter-arrival time of these decision steps, configured to approximately 100~ms (10~Hz).
At this moment, all other sensor values, such as odometry and collision flags, are sampled from the most recent buffered ROS~2 messages to construct the observation vector.
To maintain $\Delta t_{dec} \approx 100$~ms across domains without modifying the core E2E policy, the temporal synchronization strategy is adopted.
During simulation-based training, a timestamp-based blocking synchronization waits for the image timestamp to update, synchronizing the inference step with the simulator rendering speed to prevent state mismatches.
Conversely, during real-world deployment, the system operates asynchronously using a fixed 10~Hz timer callback.
At every timer event, the policy infers an action using the most recently buffered image frame, ensuring real-time responsiveness regardless of sensor or processing latency.
Despite the operational difference between simulator-driven and wall-clock-driven approaches, both modes utilize $64 \times 64$ RGB image inputs and map to the same continuous velocity commands.

\begin{table}[tb]
    \centering
    \small 
    \setlength{\tabcolsep}{3pt} 
    \caption{Specifications of ROS~2 topics and synchronization settings}
    \label{tab:ros2_specs}
    \begin{tabular}{|l|l|c|c|c|c|}
        \hline
        \textbf{Topic Name} & \textbf{Type} & \textbf{Hz} & \textbf{QoS} & \textbf{Stmp} & \textbf{Q} \\
        \hline
        
        \makecell[l]{/sensing/camera/\\front/image\_raw} & 
        \makecell[l]{sensor\_msgs/\\Image} & 
        10 & B.E. & Yes & 1 \\
        \hline
        
        \makecell[l]{/sensing/lidar/\\top/pointcloud\_raw} & 
        \makecell[l]{sensor\_msgs/\\PointCloud2} & 
        10 & B.E. & No & 1 \\
        \hline
        
        \makecell[l]{/ground\_truth/\\odom} & 
        \makecell[l]{nav\_msgs/\\Odometry} & 
        20 & B.E. & No & 1 \\
        \hline
        
        /collision\_sensor & 
        \makecell[l]{std\_msgs/\\Bool} & 
        Ev. & Rel. & No & 10 \\
        \hline
        
        \makecell[l]{/whill/controller/\\cmd\_vel} & 
        \makecell[l]{geometry\_msgs/\\Twist} & 
        10 & Rel. & No & 1 \\
        \hline
    \end{tabular}
    \vspace{1ex}
    \\
    \raggedright
    {\scriptsize Note: QoS indicates Reliability (\textit{B.E.}: Best Effort, \textit{Rel.}: Reliable). All topics use \textit{Drop Oldest} policy. \textbf{Stmp}: Timestamp dependency. \textbf{Q}: Subscriber queue size. \textbf{Ev.}: Event-driven.}
\end{table}

\section{EVALUATION}\label{sec:evaluation}
The evaluation addresses three dimensions: the reduction of the kinematic sim-to-real gap (RQ1), the reward curve of the world model-based policy for obstacle avoidance (RQ2), and direct sim-to-real deployment (RQ3).

\subsection{Experimental Setup}
Simulations and training utilize a workstation equipped with an AMD Ryzen Threadripper 7960X (24~cores), 256~GB of RAM, and an NVIDIA RTX 6000 Ada GPU.
The software stack consists of Ubuntu~22.04.5~LTS, ROS~2 Humble, AWSIM v2.0.0, and Python~3.10 with the JAX and Optuna frameworks.
For the sim-to-real transfer, the onboard computer of the physical WHILL Model CR operates on an identical Ubuntu and ROS~2 Humble environment.

\begin{table*}[t]
\centering
\caption{Comparison of metrics across different scenarios and parameters. Values represent the mean $\pm$ SD of 10 independent trials. $P$-values are relative to the initial parameters}
\label{table:evaluation_results}
\renewcommand{\arraystretch}{0.88}
\setlength{\tabcolsep}{3pt}
\footnotesize
\resizebox{\textwidth}{!}{
\begin{tabular}{|l|c|c|c|c|c|}
\hline
\textbf{Scenario} & \textbf{Parameter} & \textbf{\makecell{Trajectory RMSE \\ \small{[$m$]}}} & \textbf{\makecell{Velocity RMSE \\ \small{[$m/s$]}}} & \textbf{\makecell{Angular velocity RMSE \\ \small{[$rad/s$]}}} & \textbf{\makecell{Yaw RMSE \\ \small{[$rad$]}}} \\ \hline

\multirow{2}{*}{Straight} & Initial Values & $0.166 \pm 0.060$ & $0.060 \pm 0.004$ & $0.027 \pm 0.001$ & $0.003 \pm 0.000 $\\ \cline{2-6} 
 & Optimized Values & \makecell{$0.058 \pm 0.004$ \\ \scriptsize{($p<.001$)}} & \makecell{$0.067 \pm 0.003$\\ \scriptsize{($p<.001$)}} & \makecell{$0.028 \pm 0.001$ \\ \scriptsize{($p=.276$)}} & \makecell{$0.004 \pm 0.000$ \\ \scriptsize{($p<.001$)}} \\ \hline

\multirow{2}{*}{Turn} & Initial Values & $0.406 \pm 0.001$ & $0.178 \pm 0.000$ & $0.068 \pm 0.002$ & $0.615 \pm 0.008$ \\ \cline{2-6} 
 & Optimized Values & \makecell{$0.002 \pm 0.000$ \\ \scriptsize{($p<.001$)}} & \makecell{$0.178 \pm 0.000$ \\ \scriptsize{($p=.272$)}} & \makecell{$0.043 \pm 0.001$ \\ \scriptsize{($p<.001$)}} & \makecell{$0.064 \pm 0.022$ \\ \scriptsize{($p=.002$)}} \\ \hline

\multirow{2}{*}{Circle}  & Initial Values & $0.846 \pm 0.003$ & $0.063 \pm 0.001$ & $0.076 \pm 0.002$ & $0.547 \pm 0.005 $\\ \cline{2-6} 
 & Optimized Values & \makecell{$0.037 \pm 0.002$ \\ \scriptsize{($p<.001$)}} & \makecell{$0.037 \pm 0.002$ \\ \scriptsize{($p<.001$)}} & \makecell{$0.056 \pm 0.003$ \\ \scriptsize{($p<.001$)}} & \makecell{$0.055 \pm 0.025$ \\ \scriptsize{($p<.001$)}} \\ \hline

\multirow{2}{*}{S-curve} & Initial Values & $0.446 \pm 0.002$ & $0.044 \pm 0.002$ & $0.077 \pm 0.002$ & $0.093 \pm 0.002$\\ \cline{2-6} 
 & Optimized Values & \makecell{$0.112 \pm 0.003$ \\ \scriptsize{($p<.001$)}} & \makecell{$0.039 \pm 0.002$ \\ \scriptsize{($p<.001$)}} & \makecell{$0.069 \pm 0.002$ \\ \scriptsize{($p<.001$)}} & \makecell{$0.074 \pm 0.002$ \\ \scriptsize{($p<.001$)}} \\ \hline

 \multirow{2}{*}{\makecell{Figure-eight}} & Initial Values & $0.853 \pm 0.002$ & $0.066 \pm 0.001$ & $0.077 \pm 0.003$ & $0.189 \pm 0.004$ \\ \cline{2-6}
& Optimized Values & \makecell{$0.111 \pm 0.002$ \\ \scriptsize{($p<.001$)}} & \makecell{$0.049 \pm 0.001$ \\ \scriptsize{($p<.001$)}} & \makecell{$0.066 \pm 0.002$ \\ \scriptsize{($p<.001$)}} & \makecell{$0.118 \pm 0.004$ \\ \scriptsize{($p<.001$)}} \\ \hline

\end{tabular}
}
\end{table*}

\subsection{Reduction of Kinematic Sim-to-Real Gap}
\begin{figure}[t]
    \centering
    \begin{subfigure}{0.48\linewidth}
        \centering
        \includegraphics[width=\linewidth]{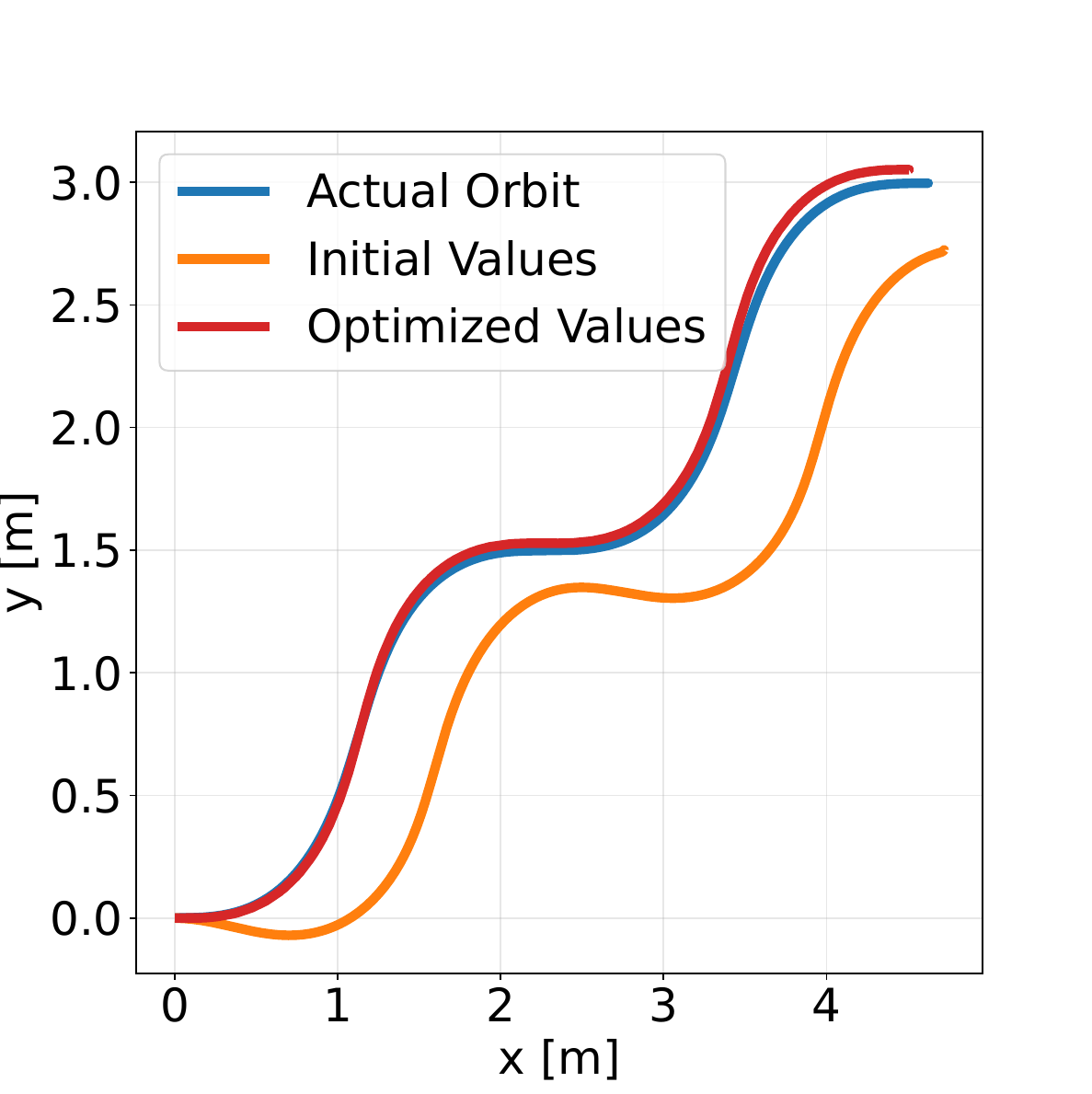}
        \caption{S-curve}
        \label{fig:rmse_s_curve}
    \end{subfigure}
    \hfill
    \begin{subfigure}{0.48\linewidth}
        \centering
        \includegraphics[width=\linewidth]{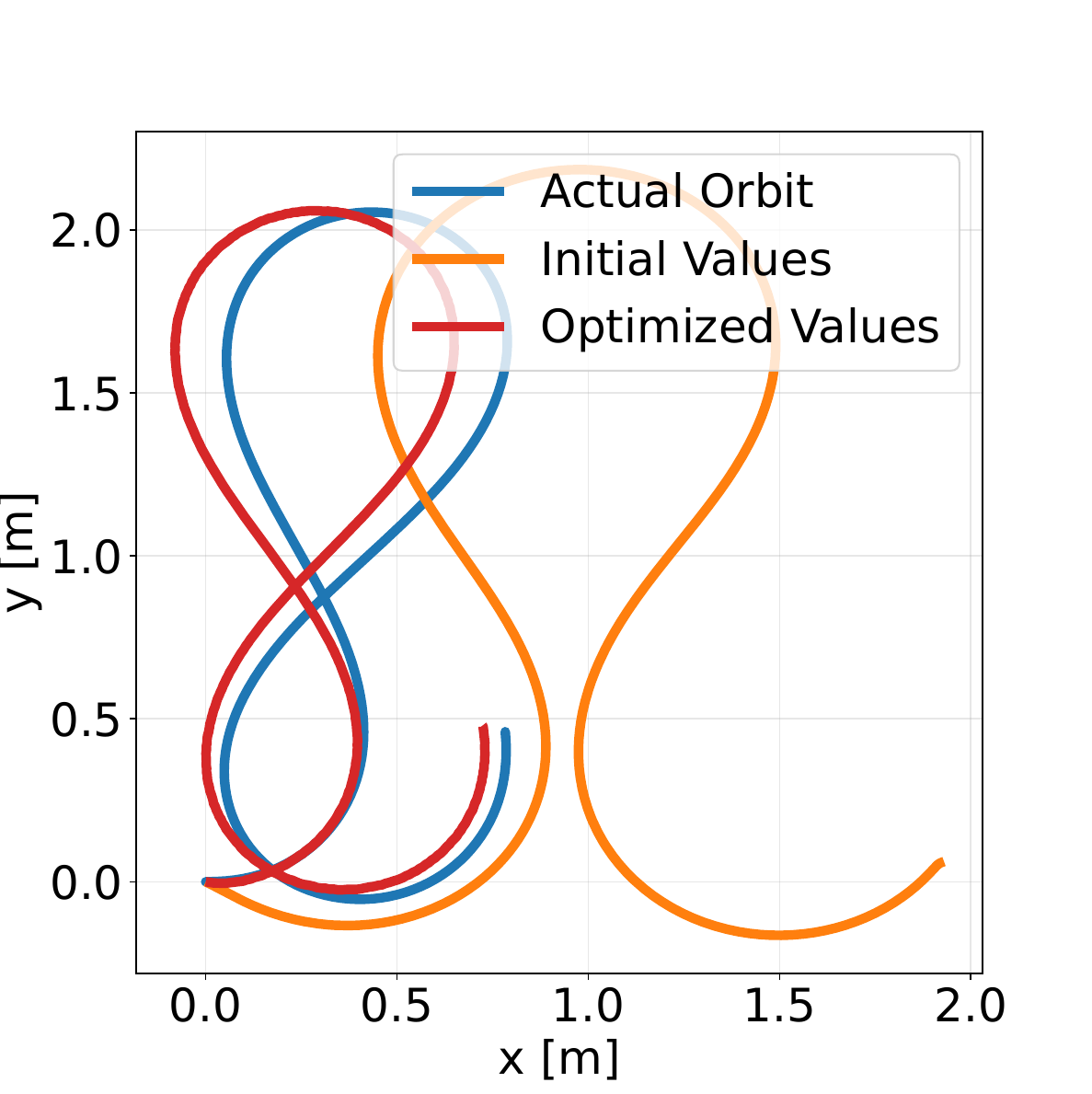}
        \caption{Figure-eight}
        \label{fig:rmse_figure_eight}
    \end{subfigure}

    \caption{Trajectories in S-curve and Figure-eight scenarios.}
    \label{fig:rmse_selected}
\end{figure}

\begin{figure}[t]
  \centering
  \begin{subfigure}[b]{0.48\linewidth}
    \centering
    \includegraphics[width=\linewidth]{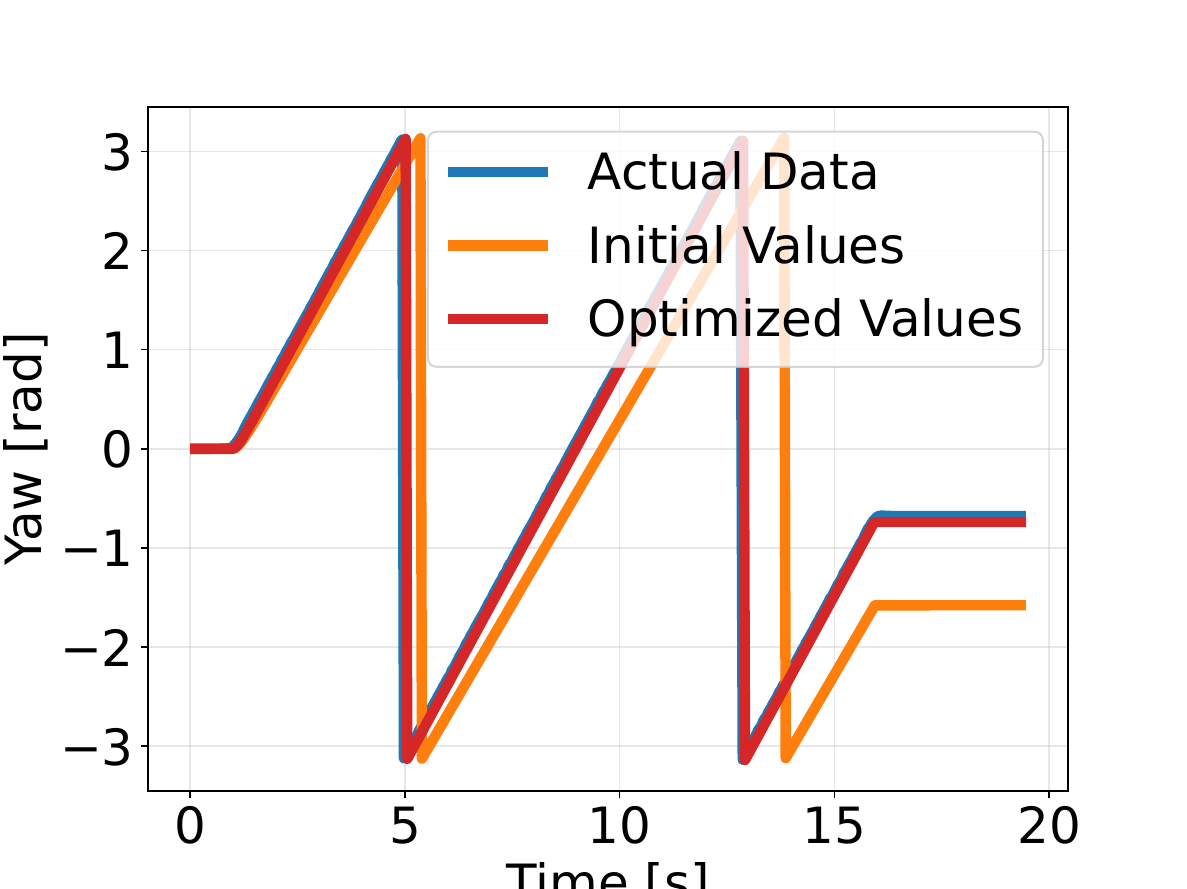}
    \caption{Turn}
    \label{fig:yaw_s_curve}
  \end{subfigure}
  \hfill 
  \begin{subfigure}[b]{0.48\linewidth}
    \centering
    \includegraphics[width=\linewidth]{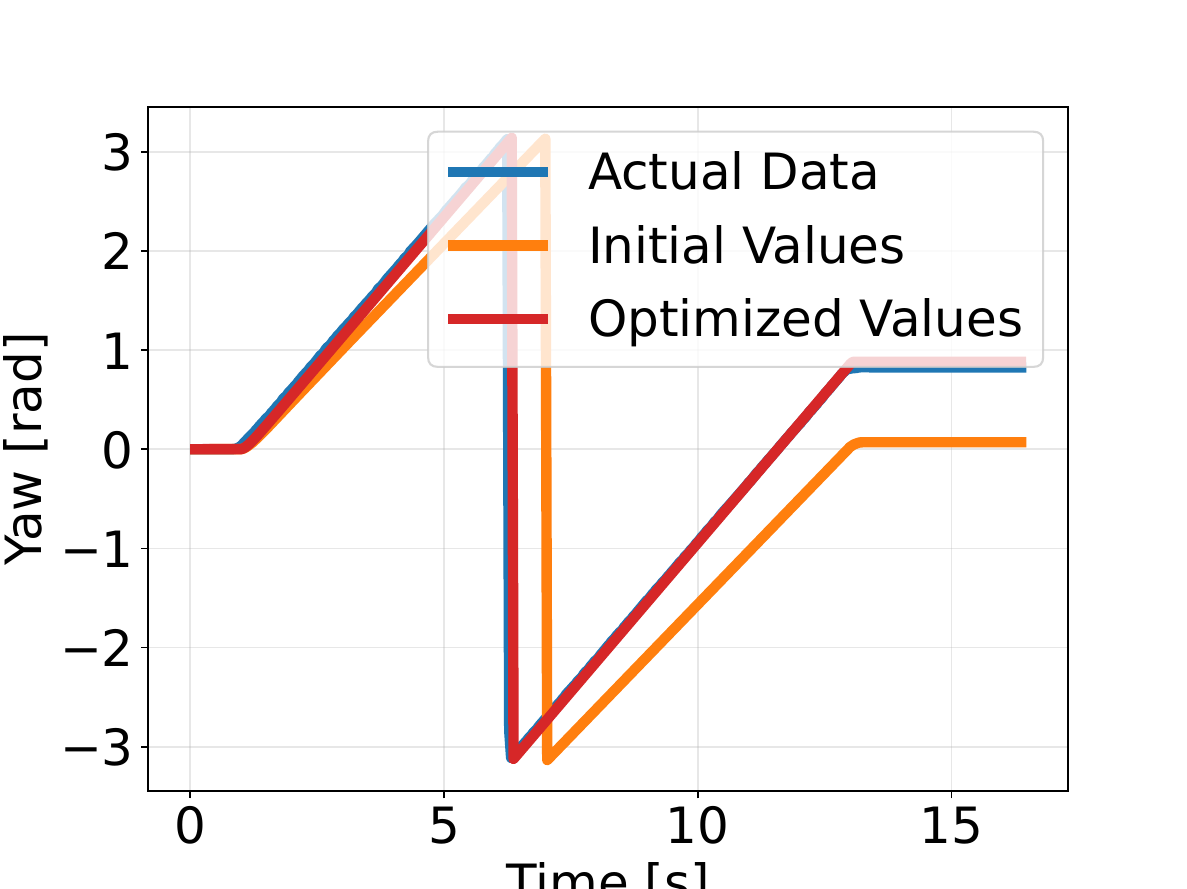}
    \caption{Circle}
    \label{fig:yaw_figure_eight}
  \end{subfigure}

  \caption{Yaw angle in Turn and Circle scenarios.}
  \label{fig:Yaw_selected}
\end{figure}

To evaluate the fidelity of the E2E learning environment (RQ1), real-world odometry and control inputs were acquired across five scenarios: Straight, Turn, Circle, S-curve, and Figure-eight.
By replaying the real-world control inputs in AWSIM (open-loop), the RMSE for trajectory, velocity, angular velocity, and yaw was calculated.
The optimized trajectories align with the real-world data, as illustrated in Fig.~\ref{fig:rmse_selected}, suppressing the outward drift seen in the initial model during complex maneuvers.
This spatial accuracy is quantitatively supported by Table~\ref{table:evaluation_results}, which shows a reduction in trajectory RMSE across the evaluated scenarios.
In the unseen Figure-eight test case, the optimization decreased the trajectory RMSE from $0.853 \pm 0.002$~m to $0.111 \pm 0.002$~m.
This result confirms that the proposed system identification generalized the physical characteristics of the vehicle without overfitting.
Furthermore, the temporal alignment of the yaw angle, illustrated in Fig.~\ref{fig:Yaw_selected}, highlights that the identified parameters compensate for the non-linear friction and lateral damping inherent to the differential drive system.
While the initial model did not capture lateral dynamics, the optimized model matched the real-world angular responses.
This improvement is validated by the yaw RMSE metrics in Table~\ref{table:evaluation_results}. For instance, the yaw error in the Turn scenario was reduced from $0.615 \pm 0.008$~rad to $0.064 \pm 0.022$~rad.
While significant improvements were observed in trajectory and yaw metrics, the RMSE for velocity and angular velocity in Table~\ref{table:evaluation_results} shows relatively minor changes between the initial and optimized models, remaining consistently low.
This result indicates that the foundational differential drive formulation in the initial model was already sufficient for tracking basic longitudinal and rotational commands.
Consequently, the optimization process isolated and corrected the non-linear lateral dynamics, such as sideslip and friction, which were the primary sources of trajectory drift, without substantially degrading the core velocity tracking performance.

\begin{figure}[t]
    \centering
    \includegraphics[width=\columnwidth]{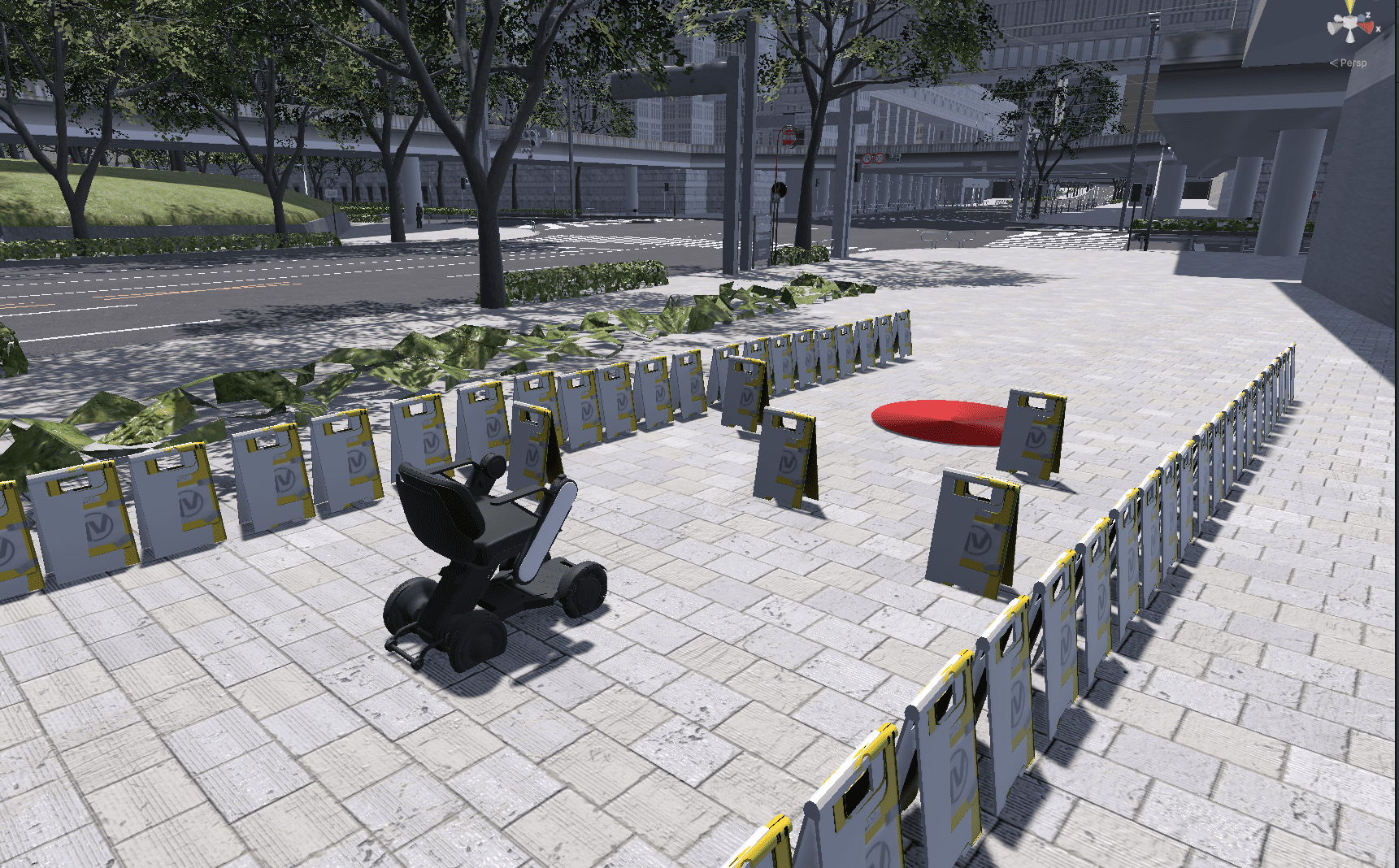}
    \caption{A scenario with obstacles along the way to the goal.}\label{scenario}
\end{figure}

\subsection{Efficacy of E2E Policy Training}
To evaluate the training platform (RQ2), an obstacle avoidance scenario is designed (Fig.~\ref{scenario}), where the agent navigates to a target using only $64 \times 64$ RGB camera images as input.
This evaluation focuses on whether the proposed AWSIM--ROS~2 environment can support the selected world model-based policy, rather than on a comprehensive algorithm benchmark.
DreamerV3 is used because its recurrent latent world model is well matched to episodic visual navigation with sparse terminal events, enabling the agent to predict task progress and collision risk from image histories while planning over continuous velocity commands.
For the simulated obstacle avoidance scenario, the predefined physical velocity limits are configured to $v_{max} = 1.0$~m/s and $\omega_{max} = 1.0$~rad/s.
Accordingly, the normalized continuous actions output by the agent are linearly scaled to a linear velocity of $v \in [-1.0, 1.0]$~m/s and an angular velocity of $\omega \in [-1.0, 1.0]$~rad/s.
The reward function is formulated to balance goal achievement, safety, and driving smoothness.
To prevent the learning signal from being corrupted by odometry drift, all distance-based rewards are computed using the ground-truth coordinates provided by the AWSIM simulator rather than estimated wheel odometry.
At each time step $t$, the total reward $R_{t}$ is defined as $R_{t} = r_{goal} + r_{dist} + r_{collision} + r_{time} + r_{smooth}$.
The goal reward $r_{goal}$ is $100.0$ if the rear axle center of the vehicle reaches within $1.0$~m of the goal center ($d_{t} \le 1.0$), and $0.0$ otherwise.
The collision penalty $r_{collision}$ is $-50.0$ if physical contact is detected, and $0.0$ otherwise.
The distance reward $r_{dist} = 30.0(d_{t-1} - d_{t})$ provides a dense progress reward proportional to the distance moved towards the goal.
Given the maximum linear velocity $v_{max}=1.0$~m/s and a control frequency of $10$~Hz, $r_{dist}$ is implicitly bounded within $[-3.0, 3.0]$ per step.
A constant time penalty $r_{time} = -0.05$ is applied per step to encourage faster task completion.
Finally, a smoothness penalty $r_{smooth} = -1.0 |\omega_{t} - \omega_{t-1}|$ is applied during active movement to suppress abrupt changes in the commanded angular velocity $\omega_{t}$ (in rad/s).

\begin{figure}[t]
    \centering
    \vspace{4pt}
    \includegraphics[width=0.9\columnwidth]{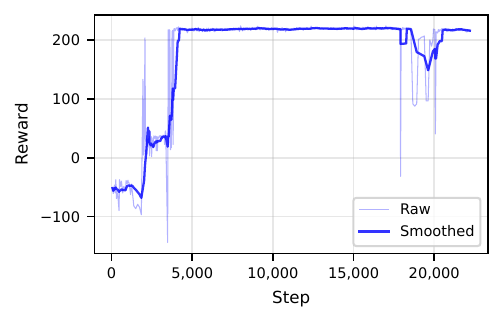}
    \vspace{4pt}
    \caption{Reward curve during training\protect\footnotemark.}\label{reward}
\end{figure}

An episode in the training and evaluation phases terminates under one of three conditions:
\begin{itemize}
\item \textbf{Success (Goal Arrival):} The episode concludes when the rear axle center of the vehicle reaches within a 1.0~m radius of the goal center.
\item \textbf{Failure (Collision):} The episode terminates immediately as a failure if the vehicle makes physical contact with an obstacle or a wall.
\item \textbf{Timeout:} To prevent infinite exploration loops, the episode is truncated if the agent fails to reach the goal within a maximum limit of 150~steps (corresponding to 15.0~seconds at 10~Hz).
\end{itemize}
The E2E policy exhibited learning progress, as shown by the reward curve in Fig.~\ref{reward}.
Following an initial exploration phase with frequent collisions, the policy demonstrated consistent goal arrivals by step 4,000.
By step 18,000, performance stabilized with an average reward of approximately 217 and an arrival time of 5~seconds.
Although a temporary performance drop occurred at step 19,000 attributable to exploration dynamics, the agent recovered and achieved task completion by step 22,000.
This performance recovery, coupled with the ability to learn obstacle avoidance from low-resolution inputs, validates the proposed environment as a practical training foundation for resource-constrained micromobility.

\footnotetext{This curve shows the learning progress of the obstacle avoidance policy for RQ2 (22,000~steps). The 200,000-step training in Section~\ref{ssub:sim2real} is conducted for a separate policy designed for the sim-to-real corridor scenario (RQ3).}

\begin{figure}[t] 
    \centering
    \begin{subfigure}[b]{0.49\linewidth}
        \centering
        \includegraphics[width=\linewidth]{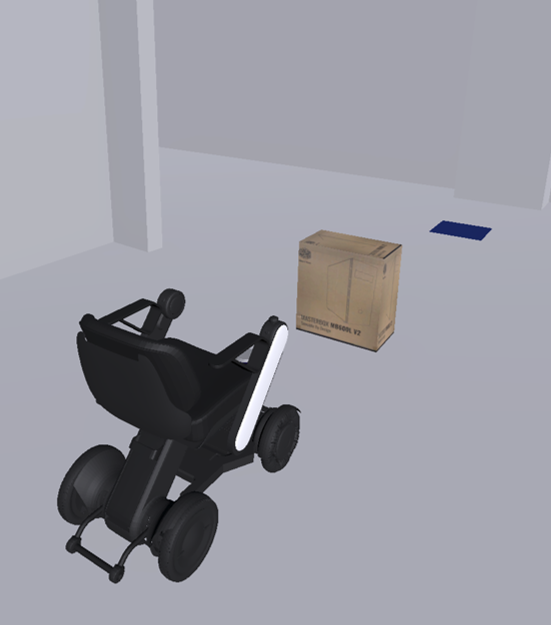}
        \caption{AWSIM.}
        \label{fig:comparison_awsim}
    \end{subfigure}
    \hfill 
    \begin{subfigure}[b]{0.49\linewidth}
        \centering
        \includegraphics[width=\linewidth]{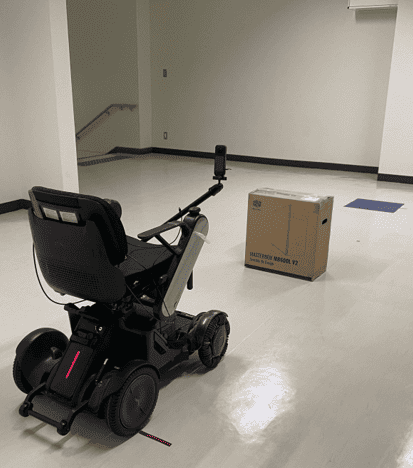}
        \caption{Real-world.}
        \label{fig:comparison_real}
    \end{subfigure}
    
    \caption{Visual comparison of the AWSIM and real-world.}
    \label{fig:sim_real_comparison}
\end{figure}

\begin{figure}[tbp]
    \centering
    \subfloat[Obstacle at 2 m]{
        \includegraphics[width=0.95\columnwidth]{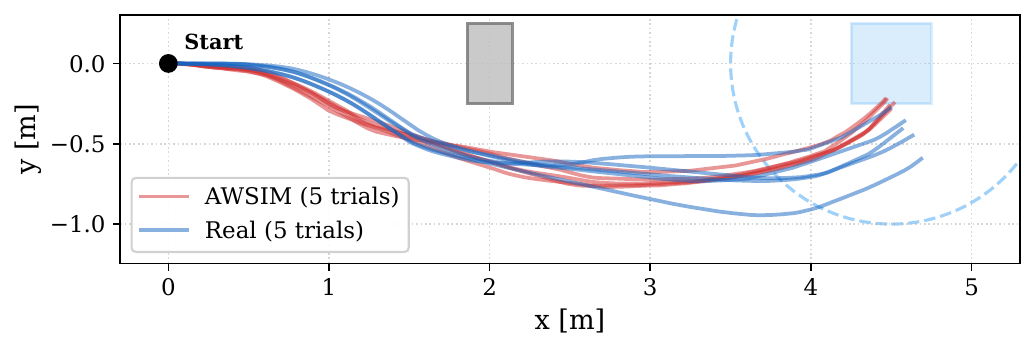}
        \label{fig:sim2real_2m}
    } \\ 
    \subfloat[Obstacle at 3 m]{
        \includegraphics[width=0.95\columnwidth]{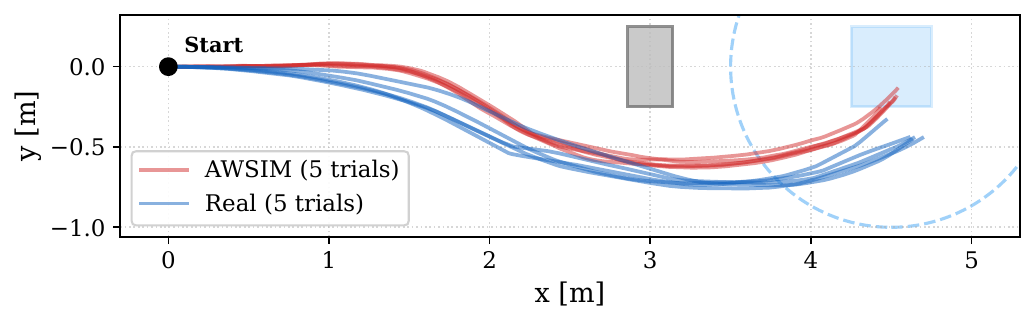}
        \label{fig:sim2real_3m}
    }
    \caption{Comparison of AWSIM and real-world trajectories during sim-to-real obstacle avoidance. The blue dotted circle indicates the 1.0 m radius goal region.}
    \label{fig:sim2real_trajectory}
\end{figure}

\subsection{Sim-to-Real Policy Transfer without Fine-Tuning}\label{ssub:sim2real}
To evaluate the real-world applicability of the learned policy (RQ3), a direct sim-to-real transfer experiment without fine-tuning was conducted using the laboratory corridor environment introduced in Section~\ref{E2E learning environment and Control}.

To establish a spatial reference, point cloud data of the physical corridor is acquired using the KISS-ICP SLAM algorithm~\cite{vizzo2023kiss} via LiDAR.
The generated 3D mesh is used to manually model a lightweight, block-based environment using Unity cube assets (Fig.~\ref{fig:sim_real_comparison}(a)).
Within this environment, a $0.5 \times 0.5$~m blue square is placed 4.5~m ahead of the vehicle as the goal, alongside a cardboard box as an obstacle, matching the physical setup (Fig.~\ref{fig:sim_real_comparison}(b)).
The E2E driving policy is trained for 200,000~steps in this AWSIM environment.
To ensure safety during physical navigation, the physical velocity limits are constrained to $v_{max} = 0.3$~m/s and $\omega_{max} = 0.6$~rad/s.
The environment wrapper linearly scales the normalized actions from the trained policy to the velocity ranges of $v \in [0.0, 0.3]$~m/s and $\omega \in [-0.6, 0.6]$~rad/s.
The training utilized the reward formulation defined in Section~IV-C, excluding the smoothness penalty ($r_{smooth}$), resulting in $R_t = r_{goal} + r_{collision} + r_{dist} + r_{time}$.
The termination conditions remain identical, except the timeout is extended to 300~steps (30~s).
Additionally, the evaluation trajectories in Fig.~\ref{fig:sim2real_trajectory} depict 20.0-second trials.

To bridge visual and spatial gaps, domain randomization (DR) is used during training.
Spatial randomization includes perturbing the initial pose of the vehicle (yaw angle between $[-15^{\circ}, 15^{\circ}]$, and lateral/longitudinal positions by $\pm 0.3$~m) and the placement of the obstacle (longitudinal translation by $\pm 1.0$~m, lateral translation by $\pm 0.5$~m).
To increase the robustness of the visual representations, data augmentation is applied to the camera inputs, randomizing brightness ($0.8$ to $1.3\times$), contrast ($0.8$ to $1.2\times$), hue ($-0.1$ to $0.1$), and saturation ($0.8$ to $1.2\times$), alongside the addition of Gaussian noise (standard deviation $0$ to $5$).
For the physical deployment, the policy weights obtained at 200,000~steps are directly transferred to the WHILL Model CR without fine-tuning.
The evaluation focuses on comparing avoidance trajectories between the simulator and the real world.
The cardboard obstacle is placed at two longitudinal distances (2~m and 3~m) ahead of the vehicle, with the goal fixed at 4.5~m.
The blue dotted circle in Fig.~\ref{fig:sim2real_trajectory} represents the 1.0~m radius goal range where the success reward triggers.
The wide acceptance radius was designed to account for the limited field of view of the camera; a tight goal region would cause the visual target to enter the frontal blind spot of the vehicle immediately before arrival, destabilizing the final approach.

As observed in Fig.~\ref{fig:sim2real_trajectory}, the physical vehicle executed collision avoidance maneuvers in all real-world trials using only the live RGB camera feed.
The trajectory data shows that the vehicle initiates evasive maneuvers at different longitudinal positions depending on the obstacle placement.
This shift in avoidance timing indicates that the policy did not overfit to a single memorized steering pattern.
Instead, the agent learned to recognize the spatial position of the obstacle from visual observations and generate avoidance commands.
While deviations between the simulated and physical trajectories exist, these deviations can be attributed to inaccuracies in the manual initial positioning.
Despite the initialization errors, the overall kinematic correlation validates the effectiveness of the system identification framework proposed in Section~\ref{sec:Design and implementation}.
Moreover, the trajectory analysis reveals a wider lateral variance in the real vehicle compared to the simulation when navigating toward the final goal. While stemming partially from residual visual domain gaps, such as lighting and material differences, this variance highlights a characteristic of the learned policy.
Instead of attempting to align with the center of the visual marker, the agent prioritizes safely clearing the obstacle to avoid collision penalties and entering the 1.0~m goal acceptance region.
The observed behavior demonstrates that the combination of safety constraints and a wide goal reward formulation absorbs visual noise and initial state errors, facilitating consistent obstacle avoidance in the physical deployment.

\subsection{Lessons Learned}
This study yields three primary findings corresponding to the reduction of the kinematic sim-to-real gap, the reward curve of the world model-based policy, and the direct physical deployment for micromobility vehicles.

RQ1: The optimization of physical parameters based on real-world driving data minimizes the sim-to-real gap in vehicle trajectories and yaw angles. The system identification framework generalized the physical characteristics without overfitting, reducing the trajectory RMSE from 0.853~m to 0.111~m in the unseen Figure-eight scenario. The optimization corrected non-linear lateral dynamics, such as sideslip and friction, while maintaining the core longitudinal velocity tracking performance.

RQ2: The E2E world model-based policy, using only RGB images, achieved curve for obstacle avoidance within the proposed environment. The policy demonstrated consistent goal arrivals by step 4,000 and reached stable performance by step 18,000 with an average reward of 217. This learning progression confirms that the proposed simulation setup provides a practical foundation for training micromobility autonomous driving systems.

RQ3: The vision-based policy trained exclusively in the simulator achieved collision-free obstacle avoidance on the physical vehicle without updating network weights. Trajectory data from the physical deployment indicates that the agent recognizes spatial obstacle positions from visual observations instead of executing memorized steering patterns. The combination of safety constraints and the goal acceptance region mitigates visual domain gaps and initial state errors during the direct sim-to-real transfer.

\section{RELATED WORK}\label{sec:Related work}

Simulation is crucial for safe autonomous driving, yet conventional car-centric platforms fail to capture the unique kinematics and shared pedestrian environments of micromobility vehicles.
A review of existing simulation platforms, E2E learning for mobile robots, and sim-to-real transfer methods positions the contributions of this study.

\subsection{Simulation Environments for Micromobility Vehicles}
Simulators such as CARLA~\cite{dosovitskiy2017carla} and AirSim~\cite{shah2017airsim} primarily target automobiles.
Recent micromobility-focused platforms, such as MetaUrban~\cite{wu2024metaurban} and URBAN-SIM~\cite{wu2025towards}, prioritize procedural generation and large-scale learning over precise kinematic reproduction.
Unlike these platforms, the proposed approach focuses on building an E2E learning environment for a specific region and vehicle model.

\subsection{E2E Learning in Mobile Robots}
E2E driving methods such as CarDreamer~\cite{gao2024cardreamer} and Follow-Me Wheelchair~\cite{salimpour2025follow}, as well as RL-based navigation algorithms (e.g., EPW-RL Navigation~\cite{chatzidimitriadis2022deep}), propose sophisticated control strategies.
However, these methods rely on simplified kinematic models in simulation, which exacerbates the dynamic discrepancy during real-world deployment.

\subsection{Sim-to-Real and System Identification}
To bridge the sim-to-real gap, approaches such as Isaac-to-Real Mobile~\cite{salimpour2025sim} utilize domain randomization~\cite{tobin2017domain}. In contrast, safely operating the WHILL Model CR demands precise system identification on the simulator side. As summarized in Table~\ref{tab:related_work_comparison}, the proposed methodology explicitly incorporates physical parameter tuning within AWSIM, establishing a foundation for real-world deployment compared to existing studies.

\begin{table}[t]
\centering
\caption{Comparison of the proposed method with related research platforms and methods}
\label{tab:related_work_comparison}
\renewcommand{\arraystretch}{1.2} 
{\tabcolsep=1.5mm
\begin{tabular}{|l|c|c|c|c|c|c|} 
\hline
Method (Reference) & SV & RM & PPT & R2 & MF & ELR \\ \hline
CARLA \cite{dosovitskiy2017carla} & & & & & & \checkmark \\ \hline
AirSim \cite{shah2017airsim} & & & & & & \checkmark \\ \hline
MetaUrban \cite{wu2024metaurban} & & & & & \checkmark & \checkmark \\ \hline
URBAN-SIM \cite{wu2025towards} & & & & \checkmark & \checkmark & \checkmark \\ \hline
CarDreamer \cite{gao2024cardreamer} & & & & & & \checkmark \\ \hline
Follow-Me Wheelchair \cite{salimpour2025follow} & \checkmark & & & \checkmark & \checkmark & \checkmark \\ \hline
EPW-RL Navigation \cite{chatzidimitriadis2022deep} & \checkmark & & & & \checkmark & \checkmark \\ \hline
Isaac-to-Real Mobile \cite{salimpour2025sim} & & & \checkmark & \checkmark & & \checkmark \\ \hline
\textbf{Proposed method} & \checkmark & \checkmark & \checkmark & \checkmark & \checkmark & \checkmark \\ \hline
\end{tabular}
}
\medskip
\begin{flushleft}
\footnotesize
 SV: Specific Vehicle, RM: Real-world Map, PPT: Physical Parameter Tuning, R2: ROS~2 Native Integration, MF: Micromobility Focus, ELR: E2E Learning Ready.
\end{flushleft}
\end{table}

\section{CONCLUSIONS}\label{sec:conclusions}

To bridge the gap between simulation and physical deployment, an E2E learning environment utilizing AWSIM and ROS~2 is developed for the kinematics of micromobility vehicles.
The optimization of physical parameters via system identification reduced the kinematic sim-to-real gap, resulting in trajectory reproduction across diverse scenarios.
Furthermore, the experimental results demonstrated that a world model-based agent (DreamerV3) trained solely within this simulated environment learned obstacle avoidance and achieved physical deployment without fine-tuning, enabling safe navigation in a physical corridor.
Consequently, the proposed environment establishes a practical and validated foundation for developing and deploying E2E driving policies in real-world micromobility applications.
Future work focuses on integrating multimodal sensors to ensure safe and efficient navigation in complex and dynamic environments.





\section*{Acknowledgment}
This work was supported by JST FOREST Grant Number JPMJFR242G.

\bibliographystyle{IEEEtran}
\bibliography{reference}

\end{document}